\documentclass[a4paper,10pt,onecolumn] {article}
\usepackage{geometry}
\usepackage{graphicx}
\usepackage{amssymb}
\usepackage{epstopdf}
\usepackage{amsmath}
\usepackage{booktabs}

\usepackage [english]{babel}
\usepackage [autostyle, english = american]{csquotes}
\MakeOuterQuote{"}

\begin{document}

\title{On-chip learning in a conventional silicon MOSFET \\ based Analog Hardware Neural Network}

\author{Nilabjo Dey$^{*,1}$,
Janak Sharda$^{*,1}$,
Utkarsh Saxena$^{1}$,\\ 
Divya Kaushik$^{1}$, 
Utkarsh Singh$^{2}$ and
Debanjan Bhowmik$^{1}$ \\
$^{*}$These authors contributed equally to the work. \\
$^{1}$Department of Electrical Engineering,\\ Indian Institute of Technology Delhi, New Delhi 110016, India, \\
$^{2}$Department of Electronics and Communications Engg., \\
 Delhi Technological University, New Delhi 110042, India \\
Email: debanjan@ee.iitd.ac.in}
\date{}
\maketitle
\begin{abstract}
On-chip learning in a crossbar array based analog hardware Neural Network (NN) has been shown to have major advantages in terms of speed and energy compared to training NN on a traditional computer.  However analog hardware NN proposals and implementations thus far have mostly involved Non Volatile Memory (NVM) devices like Resistive Random Access Memory (RRAM), Phase Change Memory (PCM), spintronic devices or floating gate transistors as synapses. Fabricating systems based on RRAM, PCM or spintronic devices need in-house laboratory facilities and cannot be done through merchant foundries, unlike conventional silicon based CMOS chips. Floating gate transistors need large voltage pulses for weight update, making on-chip learning in such systems energy inefficient.  This paper proposes and implements through SPICE simulations on-chip learning in analog hardware NN using only conventional silicon based MOSFETs (without any floating gate) as synapses. We first model the synaptic characteristic of our single transistor synapse using SPICE circuit simulator and benchmark it against experimentally obtained current-voltage characteristics of a transistor. Next we design a Fully Connected Neural Network (FCNN) crossbar array using such transistor synapses.  We also design analog peripheral circuits for neuron and synaptic weight update calculation, needed for on-chip learning, again using conventional transistors.  Simulating the entire system on SPICE circuit simulator, we obtain high training and test accuracy on the standard Fisher's Iris dataset, widely used in machine learning. We also account for device variability and noise in the circuit, and show that our circuit still trains on the given dataset. We also compare the speed and energy performance of our transistor based implementation of analog hardware NN with some previous implementations of NN with NVM devices and show comparable performance with respect to on-chip learning. Easy method of fabrication makes hardware NN using our proposed conventional silicon MOSFET really attractive for future implementations.  
\end{abstract}

\twocolumn

\sloppy

\section{Introduction}
Neural network (NN) algorithms are being widely used by the machine learning and data sciences community currently to solve several data classification and regression tasks \cite{LeCun}. However implementing NN on a traditional computer built on von Neumann architecture (memory and computing physically separated) involves continuous transfer of information between the memory and computing units. This von Neumann bottleneck leads to lower performance in terms of speed and energy consumption \cite{NeuralNetSurvey_Misra, NeuralNetSurvey_Schuman, BoahenStanfordReview,GeffBurrJPhysDReview,PCMReview_AbuSebastian,Kaushik_IEEEReview}. Hence researchers have come up with specialized hardware NN implementations to get rid of the von Neumann bottleneck  \cite{IBMTrueNorth,SpinNaker,Lohi_Intel,NeuralNetComparison_Diamond}.  Among these implementations, analog hardware NN uses a crossbar array of synaptic devices to perform computing at the location of the data itself  \cite{GeffBurrJPhysDReview, PCMReview_AbuSebastian, GeffBurrIEDM2015}. The fact that such crossbar array enables execution of  Vector Matrix Multiplication (VMM), inherent in a FCNN algorithm, in a parallel fashion makes it suitable both for forward inference \cite{GeffBurrJPhysDReview} or on-chip learning (training in hardware). In fact, on-chip learning in such crossbar array has been considered to be faster and more energy efficient than conventional training of NN on GPU \cite{GeffBurrIEDM2015}.

A synaptic device in a crossbar array based analog hardware NN must have several conductance states that can be controlled electrically to store and update the weight values of the NN. Several Non Volatile Memory (NVM) devices like floating gate transistors, chalcogenide based Phase Change Memory (PCM) devices, oxide based Resistive Random Access Memory (RRAM) devices and spintronic devices  have been proposed and used as synaptic devices in previous implementations of analog hardware NN \cite{GeffBurrJPhysDReview,PCMReview_AbuSebastian, Kaushik_IEEEReview, NeuromorphicReview_PhilipWong, MSuriBook, GrollierReview, PCMNature, memristorNature, PCMManan, FloatingGateX}.
However, these NVM devices have several issues associated with them with respect to achieving on-chip learning in crossbar arrays made of them. Floating gate transistor synapses need high voltage pulses for weight update and have low endurance  \cite{FloatingGateX, FloatingGate1, FloatingGate2,FloatingGate3,FloatingGateDisadvantages,FloatingGateDisadvantages2}. Memristive oxide based Resistive Random Access Memory (RRAM) devices and Phase Change Memory (PCM) devices \cite{PCMNature, memristorNature,PCMManan} exhibit an asymmetric/uni-polar and non-linear dependence between conductance (and weight) update and programming pulses, which affects the accuracy during on-chip learning of crossbar arrays that use such devices \cite{GeffBurrJPhysDReview,GeffBurrTED,ShimengIEDM,RPUFrontNeuroscience}. Moreover, for fabricating  RRAM, PCM or spintronic device based hardware NN systems \cite{GeffBurrJPhysDReview,Kaushik_IEEEReview,SpintronicRoadmap}, dedicated in-house fabrication facilities are needed since they involve novel materials. The system cannot be designed in-house and then fabricated elsewhere e.g. in commercial merchant foundries, unlike silicon based conventional CMOS circuits and systems.

Instead if conventional Metal Oxide Semiconductor Field Effect Transistors (MOSFETs) with silicon (Si) as semiconducting material and SiO$_2$ as gate oxide could be used as synaptic elements and analog values of weight could be stored in them, unlike what's done in an SRAM cell where digital bits are stored, then fabrication of analog hardware NN will be much easier. In this paper, we propose such a conventional MOSFET as a three terminal synaptic element in analog hardware NN. In Section 2, we show through SPICE simulations at 65 nm technology node on Cadence Virtuoso circuit simulator that the conductance between drain and source of the transistor (first and second terminal) can be controlled between several analog values, which represent the weight values, by changing the voltage applied at the gate of the transistor  (third terminal) (Fig. ~\ref{Synapse1}). We benchmark our data against experimentally measured data on a n-MOSFET present inside a commercially available chip (Fig. ~\ref{Synapse2})  Gate voltage can be changed by applying current pulses at the gate and charging/ discharging the gate oxide, as we show in our simulations (Fig. ~\ref{Synapse3}). Conductance (weight) vs programming gate current pulse plot is found to be fairly linear and symmetric for positive and negative weight update for our proposed transistor synapse (Fig. ~\ref{Synapse3}), unlike PCM and RRAM based synaptic devices in which non-linearity and asymmetry/ unidirectionality in conductance response degrades the overall neural network performance \cite{GeffBurrJPhysDReview,GeffBurrTED,ShimengIEDM,RPUFrontNeuroscience}.  

In Section 3 we design a Fully Connected Neural Network (FCNN) circuit in crossbar topology using such synaptic transistors at 65 nm technology node that carries out the input Vector- weight Matrix Multiplication (VMM), characteristic of FCNN (Fig. ~\ref{Network1}). We design, using transistor and transistor based op-amps, analog neuron circuit (Figure ~\ref{Neuron Circuit}) and Stochastic Gradient Descent (SGD) algorithm  \cite{LeCun} based synaptic weight update calculation circuit (Figure ~\ref{SGD}). This weight update circuit sends current pulses to the gates of the synaptic transistors. These gate current pulses update the weight values of the FCNN (Fig.  ~\ref{Synapse3},~\ref{Network1}).

In Section 4, using SPICE simulations of the whole system on Cadence Virtuso circuit simulator, we demonstrate on-chip learning in our proposed hardware  (Fig.~\ref{Schematic}) on the Fisher's Iris dataset- a popular machine learning dataset (Fig. ~\ref{output}) \cite{FisherIris}. In Section 5, we next compare the speed and energy performance of our transistor synapse based analog NN against analog implementations of the same through previously proposed NVM devices with respect to on-chip learning on the exact same dataset \cite{ memristorNature,GeffBurrTED,   Saxena, Bhowmik_JMMM, Kaushik_BioCAS}. Speed and energy consumption are almost equal for our proposed transistor synapse and spintronic (domain wall based) synapse (Table I). Compared to RRAM synapse, speed for transistor synapse is much higher. Also energy is several orders lower (Table I), again because of the asymmetric nature of conductance response of RRAM synapse as opposed to the synaptic nature of the same in transistor synapse. In Section 6, we argue that our proposed transistor synapse based FCNN circuit trains itself even in the presence of synaptic device variability and noise in input voltages.

 Thus we show that Si MOSFET synapse can be considered as an attractive candidate for implementation of hardware analog NN.  To the best of our knowledge, this is the first proposal and demonstration through simulations of on-chip learning on analog hardware NN using only conventional Si MOSFETs  as synapses. It is to be noted that \cite{FloatingGateX,FloatingGate1,FloatingGate2,FloatingGate3} propose synaptic behaviour of MOSFET by using a floating gate. However weight modulation in floating gate synapse is much slower because electrons have to be injected inside the gate through a tunneling mechanism for a change of weight. Also, the voltages needed for such weight update are very large \cite{FloatingGateDisadvantages}.Number of times charge can be injected into/ ejected out of the floating gate is also limited, leading to low endurance \cite{FloatingGateDisadvantages2}. Since weight has to be frequently updated for on-chip learning, floating gate transistor hence doesn't make for a good synapse.   On the other hand, our synaptic MOSFET is conventional, doesn't have a floating gate and hence doesn't suffer from those disadvantages.
 
It is also to be noted that earlier reports of conventional silicon transistor based synapse use multiple transistors to store each bit of the weight value stored in the synapse \cite{ChetanThakur1,ChetanThakur2}. On the other hand, analog values of weight are stored in a single transistor in our proposed synapse as different conductance states.

Also, in this paper, we demonstrate the capability of our transistor synapse in a non-spiking network, trained "on-chip" through the much developed Stochastic Gradient Descent (SGD) algorithm benchmarked on several machine learning datasets, as opposed to spiking network which mostly uses Spike Time Dependent Plasticity (STDP) enabled training algorithm \cite{GeffBurrTED,Diehl1,BiPoo,SpikingNetReview,Kaushik_STDP1,SpikingNetManan}. Convergence properties and highly accurate training results have not been demonstrated on large datasets in STDP enabled spiking network algorithms  to the extent they have been in SGD based non-spiking network algorithms  \cite{SpikingNetReview,GeoffreyBurrIBMInternalReport}.

\section{Characterization of a single conventional MOSFET as a synaptic device}

\begin{figure*}[!t]
\centering
\includegraphics[width=0.8\textwidth]{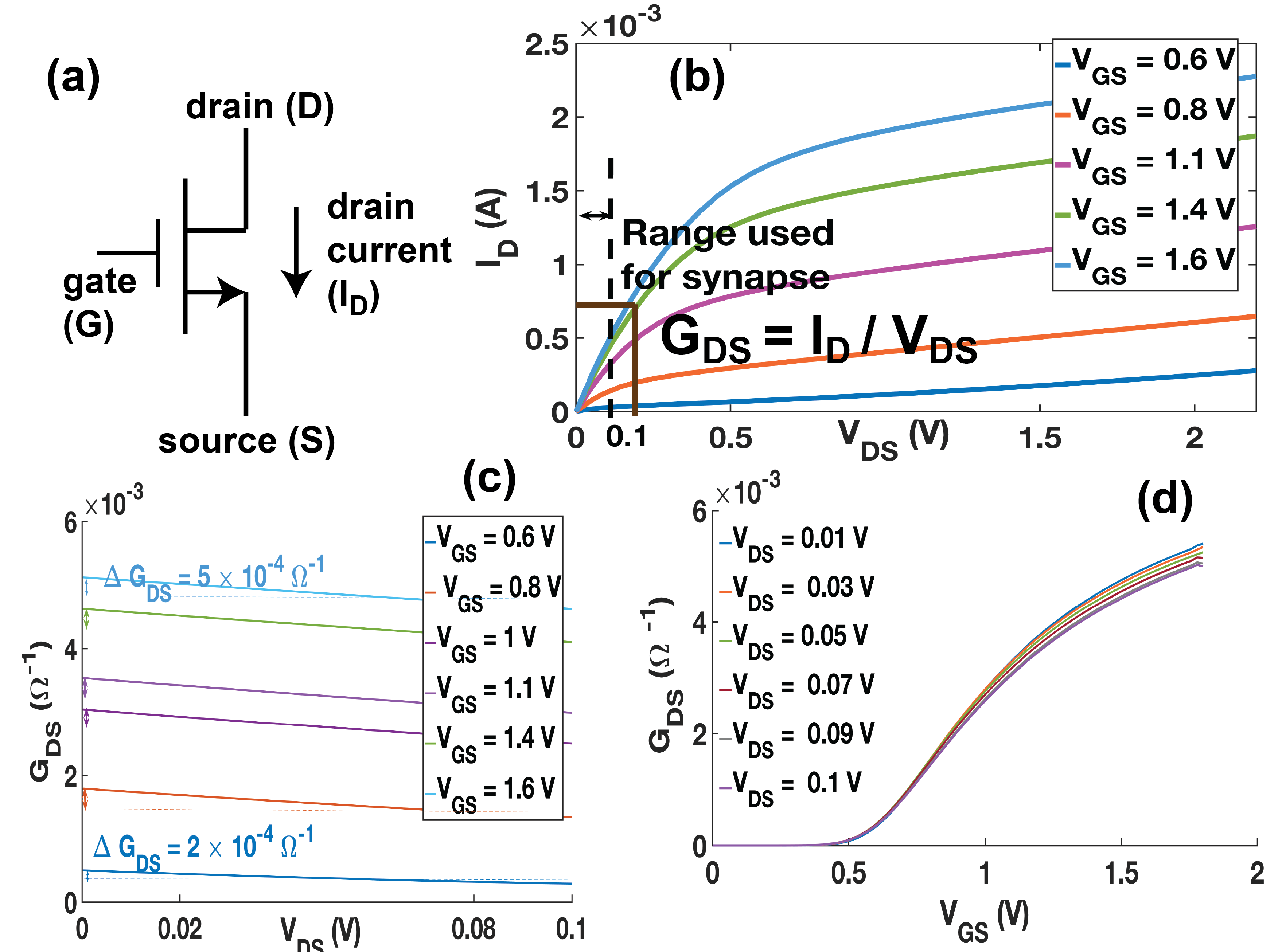}
\makeatletter
\caption{(a) Schematic of proposed n-MOSFET synapse. Following sub-plots are obtained from simulation of the MOSFET at 65 nm node on SPICE circuit simulator. (b) Drain current ($I_{D}$) as a function of drain to source voltage ($V_{DS}$ for different applied gate to source voltages ($V_{GS}$) (c) Conductance between drain (D) and source (S) ($G_{DS}$) plotted as a function of  $V_{DS}$ for different values of $V_{GS}$. (d) ($G_{DS}$) plotted as a function of  $V_{GS}$ for different values of $V_{DS}$.}
\label{Synapse1}
\end{figure*}

\begin{figure*}[!t]
\centering
\includegraphics[width=0.8\textwidth]{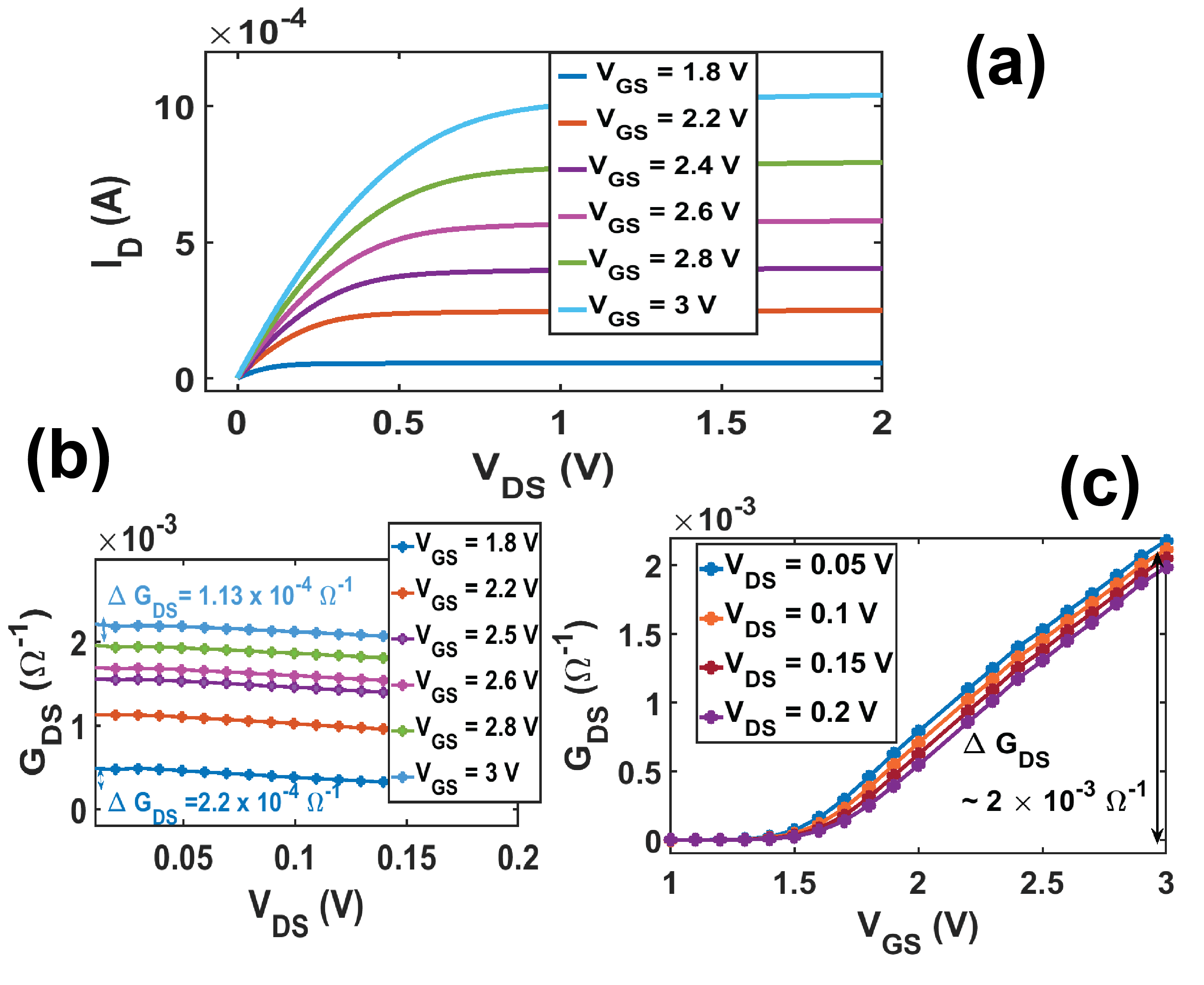}
\makeatletter
\caption{Experimental data obtained from measurement on single n-MOSFET in commercially available CD4007 chip (a) Drain current ($I_{D}$) as a function of drain to source voltage ($V_{DS}$ for different applied gate to source voltages ($V_{GS}$) (b) Conductance between drain (D) and source (S) ($G_{DS}$) plotted as a function of  $V_{DS}$ for different values of $V_{GS}$. (c) ($G_{DS}$) plotted as a function of  $V_{GS}$ for different values of $V_{DS}$.  }
\label{Synapse2}
\end{figure*}

Fig. ~\ref{Synapse1}(a) shows schematic of a Si-SiO$_2$ based n-MOSFET we propose as synapse in this paper. We simulate it at the 65 nm technology node through SPICE simulations on Cadence Virtuoso using the United Microelectronics Corporation (UMC) library. Drain current ($I_{D}$) vs drain to source voltage ($V_{DS}$) characteristic is linear for a certain range of $V_{DS}$ (0 - 0.1 V in this case) for gate to source voltage $V_{GS}$ in between 0.6 V and 1.6 V (Fig. ~\ref{Synapse1}(b)). This behavior is expected for conventional MOSFET \cite{ChenmingHu} and matches qualitatively with $I_{D}$- $V_{DS}$ characteristic we experimentally measure on a single n-MOSFET present inside the commercially available CD4007 inverter chip and accessible through package terminals (Fig. ~\ref{Synapse2}(a)). We operate $V_{DS}$ between 0 and 0.1 V, and  $V_{GS}$ between 0.6 and 1.6 V for functioning as synapse throughout this paper.

For any combination of $V_{GS}$ and $V_{DS}$, ratio of $I_{D}$ to $V_{DS}$ determines drain to source conductance ($G_{DS}$), which is a function of both $V_{GS}$ and $V_{DS}$ ($G_{DS}(V_{GS},V_{DS})$)   (Fig. ~\ref{Synapse1}(b)). We observe that for a fixed $V_{GS}$, change in $G_{DS}$ ($\Delta G_{DS}$) is in the order of $10^{-4}$ $\Omega^{-1}$ when $V_{DS}$ varies in the full range we have selected (0 - 0.1 V) (Fig. ~\ref{Synapse1}(c)). However, for a fixed $V_{DS}$ when  $V_{GS}$ varies in full range (0.6 - 1.6 V), change in $G_{DS}$ is $~ 5 \times 10^{-3} \Omega^{-1}$ , which is one order higher than change in $G_{DS}$ due to full sweep of $V_{DS}$. Thus $G_{DS}$ can just be approximated as a function of $V_{GS}$ and not $V_{DS}$ (and in extension $I_{DS}$)  in the selected range of operation. 

Hence we can write 
\begin{equation}
I_{D}= G_{DS}(V_{GS},V_{DS})V_{DS} \approx G_{DS}(V_{GS})V_{DS}  
\end{equation}

\begin{figure*}[!t]
  \centering
  \includegraphics[width=0.55\textwidth]{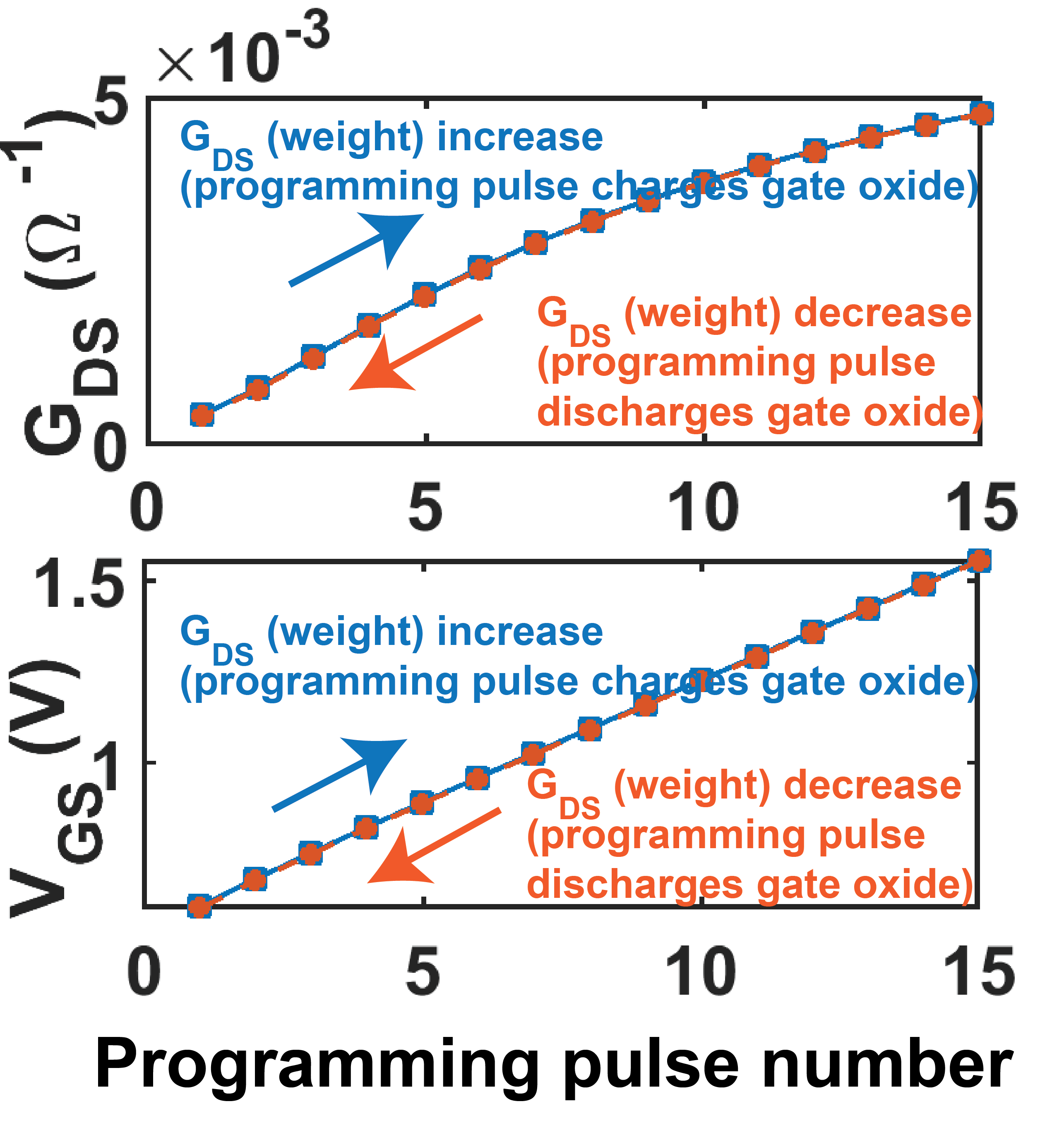}
  \makeatletter
  \caption{Conductance $G_{DS}$ (and hence weight)and corresponding gate voltage ($V_{GS}$) increase linearly due to programming current pulses which charge the capacitive gate oxide of the MOSFET synapse (blue plot). Conductance and gate voltage decrease linearly due to programming current pulses of opposite polarity which discharge the gate oxide (orange plot) }
  \label{Synapse3}
\end{figure*}

\begin{figure*}[!t]
  \centering
  \includegraphics[width=0.5\textwidth]{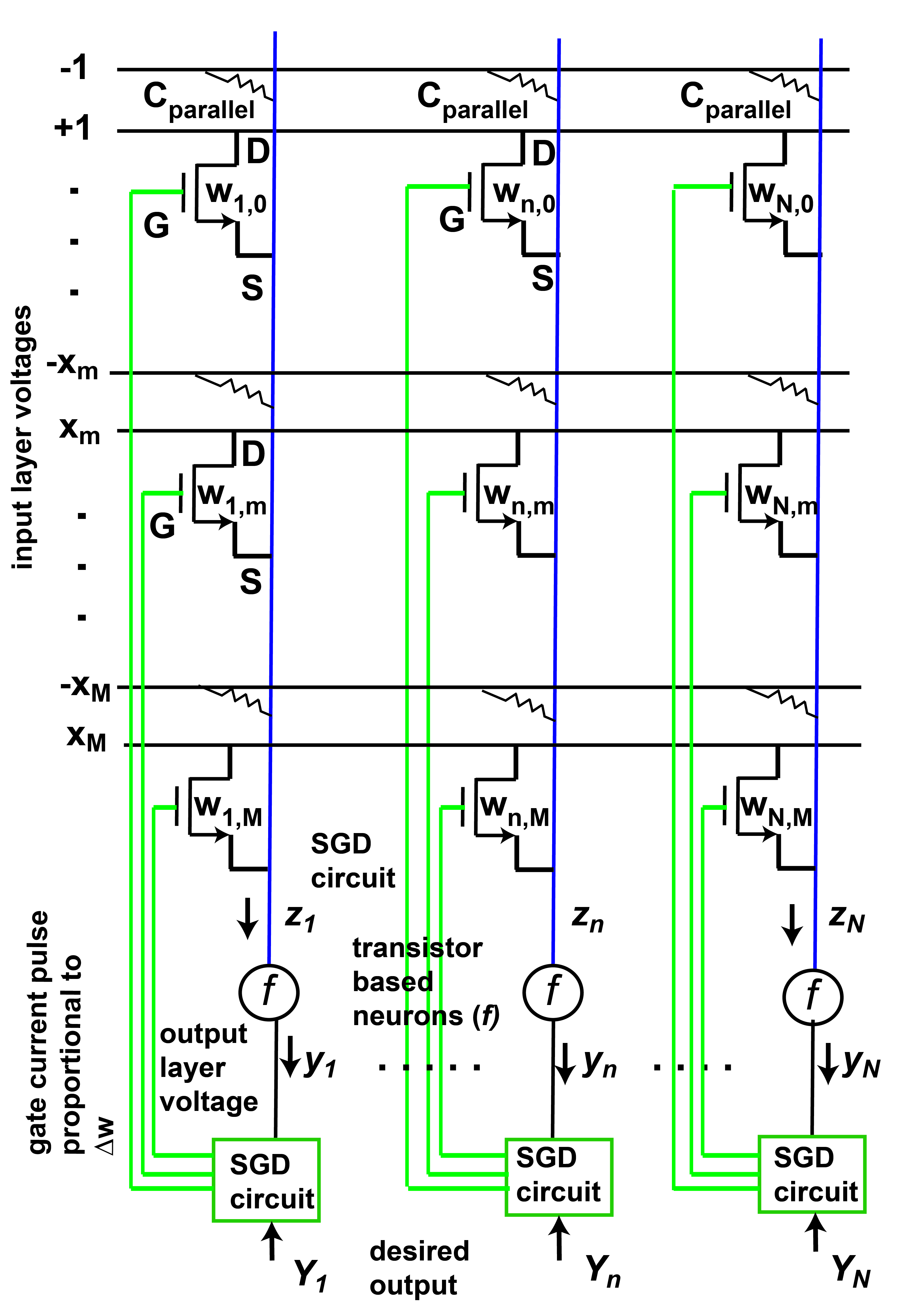}
  \makeatletter
  \caption{Schematic of transistor synapse based cross-bar architecture for weight matrix-input vector multiplication, followed by transistor based neuron circuit for activation function (f) (shown in Figure ~\ref{Neuron Circuit}) and then feedback circuit that evaluates change in weight needed based on output of neurons and desired output using Stochastic Gradient Descent (SGD) method (shown in Figure ~\ref{SGD}). The feedback circuit then applies current pulses at the gates of the transistor synapses such that the gate voltage, and hence the weight stored by the transistor synapse, changes by the desired amount}
  \label{Network1}
\end{figure*}

\begin{figure*}[!t]
  \centering
  \includegraphics[width=0.6\textwidth]{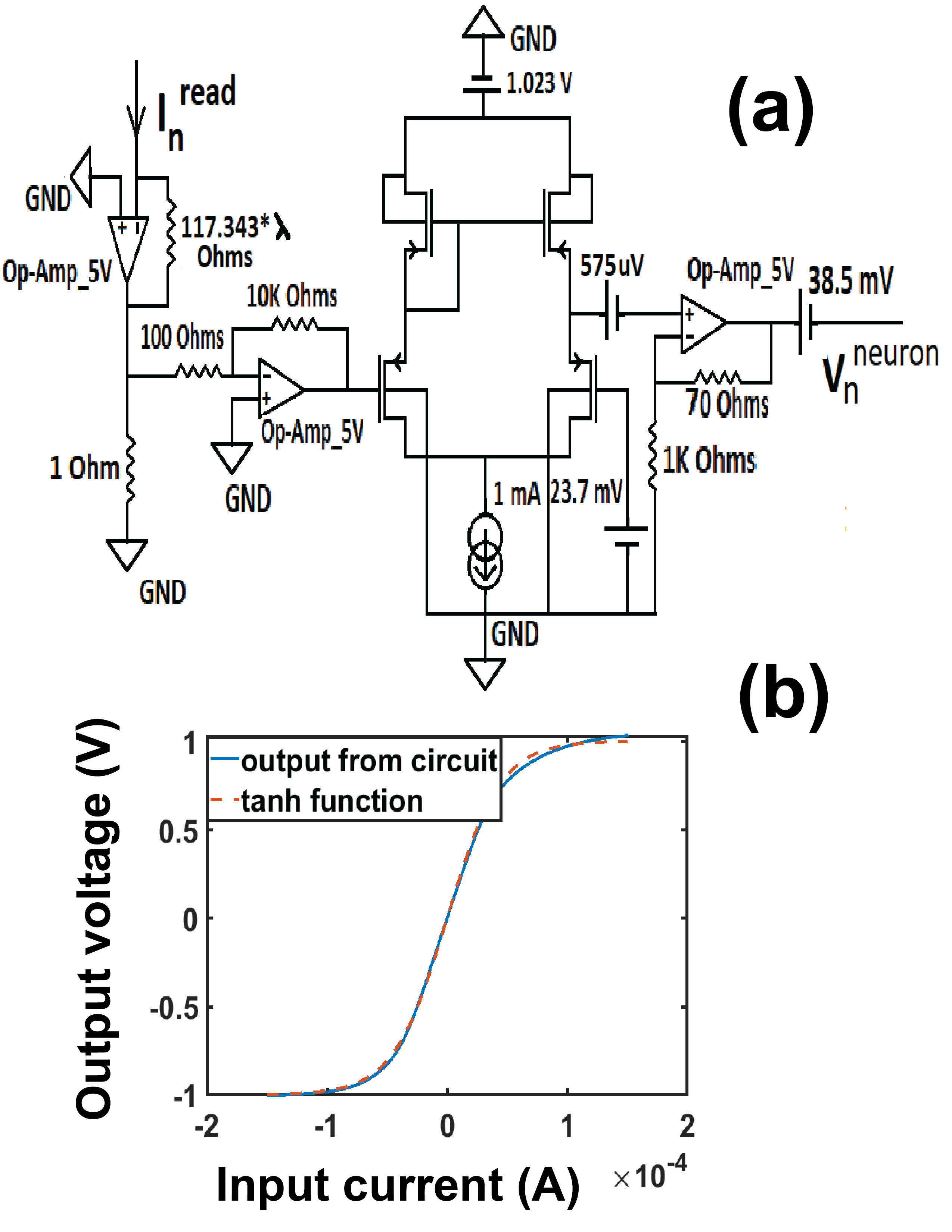}
  \makeatletter
  \caption{(a) Neuron circuit present at each output node of the crossbar circuit in Figure ~\ref{Network1} shown in details here. (b) Output voltage of neuron circuit as a function of input current , obtained from SPICE simulation is shown with solid line (blue). Ideal tanh function is plotted with dashed line (red)}
  \label{Neuron Circuit}
 \end{figure*}

Experimentally measured data on transistor in CD4007 chip qualitatively matches with this observation  (Fig. ~\ref{Synapse2}(b),(c)). For a quantitative match, the specifications of the simulated and experimentally measured transistors need to match which has not been the case in this work. When our simulated transistor is used as synapse in $M$ input nodes $\times$ $N$ output nodes crossbar array based analog hardware FCNN as shown in Fig.~\ref{Network1}(a),  input vector- weight matrix multiplication, or VMM operation, takes place as a part of the feedforward computation both for the training phase and testing/ inference phase \cite{GeffBurrJPhysDReview}. During this process, the input vector ${\{x_1,x_2,x_3...x_m..x_{M}\}}$ corresponding to a training sample acts as drain voltages on the transistor synapses as shown in Fig.~\ref{Network1}(a). The sources of all the transistor synapses are maintained at 0 V using op-amps at the input stage of the neuron circuits we design (Fig. ~\ref{Neuron Circuit}). Thus for a transistor synapse connecting input node $m$ with output node $n$, its $V_{DS}$ is proportional to input $x_{m}$. If its conductance $G_{DS}$ represents its weight $w_{n,m}$ then from equation (1) its drain current $I_{D}$ turns out to be proportional to $w_{n,m}x_{m}$. Even when $V_{DS}$ and hence $I_{D}$ changes during the VMM operation in both training and testing phase of the hardware, as long as $V_{DS}$ stays in the chosen range of 0 and 0.1 V this relation holds true. In any case, since we show training through SPICE simulations of the entire hardware in this paper, any effect of higher order term of drain voltage $V_{DS}$ on the current $I_{D}$ is taken into account in our analysis and final accuracy results.

Conductance ($G_{DS}$) and hence synaptic weight update is carried out by applying current pulses at the gate of the MOSFET synapse and charging/ discharging the gate oxide. Fig.~\ref{Synapse3} shows that when pulses of the same current magnitude (132 nA) and same duration (1 ns) with polarity such that gate oxide charges up, gate voltage $V_{GS}$ and hence $G_{DS}$ increases linearly with pulse number (blue plot). Similarly starting from the highest $G_{DS}$, programming pulses of opposite polarity but of same magnitude and duration discharge the oxide and $G_{DS}$ decreases (orange plot). This linear and symmertrical/bipolar nature of weight update with programming pulse number is not the case in RRAM and PCM based synaptic devices, making implementation of on-chip learning in crossbar arrays made of such synapses quite challenging \cite{GeffBurrJPhysDReview,GeffBurrTED,ShimengIEDM}. Even when a pair of such RRAM or PCM devices is used as a single synapse as has been done before, frequent RESET pulses, which are of long time duration and consume a lot of energy, are still needed to carry out the weight update scheme \cite{PCMManan,GeffBurrTED}.

The weight update process for our proposed transistor synapse is repeated for every training sample in each epoch to obtain high training accuracy after a certain number of epochs and thus achieve on-chip learning, as we show next.

\section{Design of crossbar Fully Connected Neural Network (FCNN) circuit and feedback circuit to train it}

Analog crossbar array of our proposed synaptic transistors is designed in Cadence Virtuoso SPICE circuit simulator  next to implement a Fully Connected Neural Network (Fig. ~\ref{Network1}) \cite{LeCun,Haykin}.  If the input layer has $M$ nodes and output layer has $N$ nodes then for an input vector ${\{x_1,x_2,x_3...x_m..x_{M}\}}$ at the input layer the output vector at the output layer is given by ${\{y_1,y_2,y_3...y_n..y_{N}\}}$ where:
\begin{equation}
y_n=f(z_n) = \frac{2}{1+e^{-\lambda z_n}}-1
\end{equation}
and 
\begin{multline}
z_n = w_{n,1}x_1 + w_{n,2}x_2 + .... \\
+w_{n,m}x_m+.. w_{n,M}x_{M}+w_{n,0}\\
=(\Sigma_{m=1}^{m=M}w_{n,m}x_m)+ w_{n,0}
\end{multline}
where $w_{n,m}$ is the weight of the synapse connecting input node $m$ with output node $n$, $f$ is the tanh activation function and $\lambda$ is a parameter in the function. As discussed in the previous section, to implement the weight matrix- input vector multiplication of equation (3) in hardware, we map the weight to the conductance between drain and source of MOSFET synapse ($G_{DS}$). Minimum value of weight corresponds to minimum conductance value and maximum value of weight corresponds to maximum conductance value within the chosen gate voltage range where conductance is almost linearly proportional to gate voltage (Fig.~\ref{Synapse2}(c)). Voltage proportional to input at node $m$ ($x_m$) is applied at the drain ($V_{DS}$) of transistor synapse connecting input node $m$ with output node $n$ (Fig. ~\ref{Network1}), in the form of a 1ns long pulse corresponding to one training sample. $V_{DS}$ corresponding to maximum value of $x_m$ is 0.1 V. $G_{DS}$   varies little with $V_{DS}$ when $V_{DS}$ is below 0.1 V as seen in our simulations (Fig.~\ref{Synapse2}(b)).

\begin{figure*}[!t]
  \centering
  \includegraphics[width=0.7\textwidth]{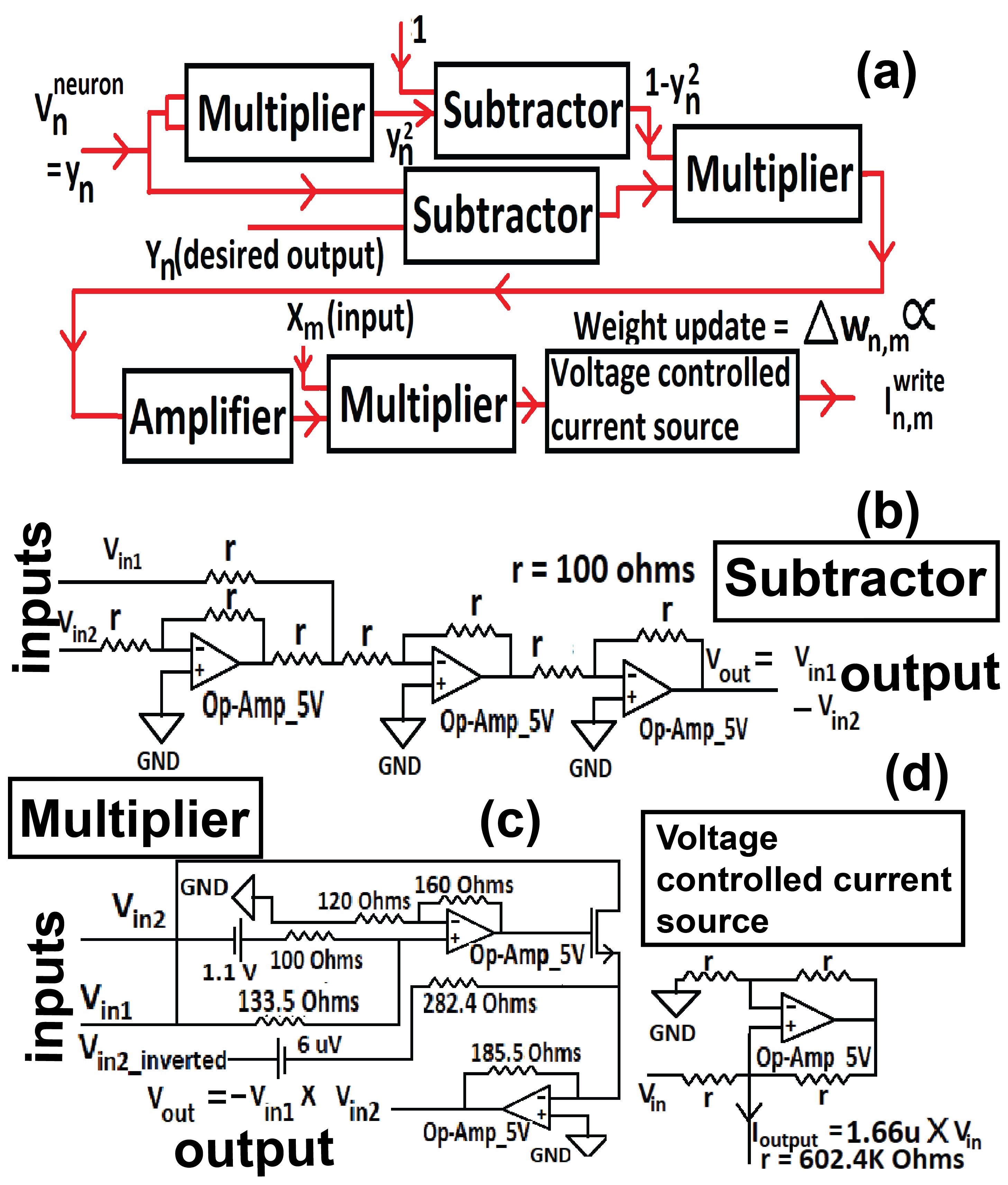}  
    \makeatletter
\caption{(a) Analog peripheral feedback circuit we designed, using conventional silicon transistors and op-amps made of them, on circuit simulator for weight update following SGD algorithm (b) Design of subtractor block (c) Design of multiplier block (d) Design of voltage controlled current source circuit.} 
\label{SGD}
\end{figure*}

Since weight takes both positive and negative values while conductance is only positive, a resistance is added in parallel to a transistor synapse and voltage proportional to negative of $x_m$ is applied to it (Fig. ~\ref{Network1}). Thus current proportional to  $w_{n,m}x_m$ flows between drain and source of the transistor. Currents of all transistors connected to output node $n$ add up following Kirchoff's Current law generating current proportional to $z_n$ in equation (3).

Current proportional to $z_n$ next enters a transistor based analog neuron circuit that executes tanh activation function ($f$) of equation (2) \cite{NeuronCircuit1,NeuronCircuit2}. Thus output voltage of neuron circuit corresponds to $y_n$ at output node n (Fig. ~\ref{Neuron Circuit}). We design this neuron circuit using a set of pre-amplifiers composed of transistor based op-amps and a transistor based differential amplifier circuit, again on Cadence Virtuoso simulator (Fig. ~\ref{Neuron Circuit}(a)). Output voltage vs input current plot (Fig. ~\ref{Neuron Circuit}(b)) for the neuron circuit shows desired tanh behavior of equation (1).  Hyperparameter ($\lambda$) in the function (equation 2) can be adjusted by changing a resistance in the neuron circuit as shown in Fig.~\ref{Neuron Circuit}(a)).

\begin{figure*}[!t]
  \centering
  \includegraphics[width=0.6\textwidth]{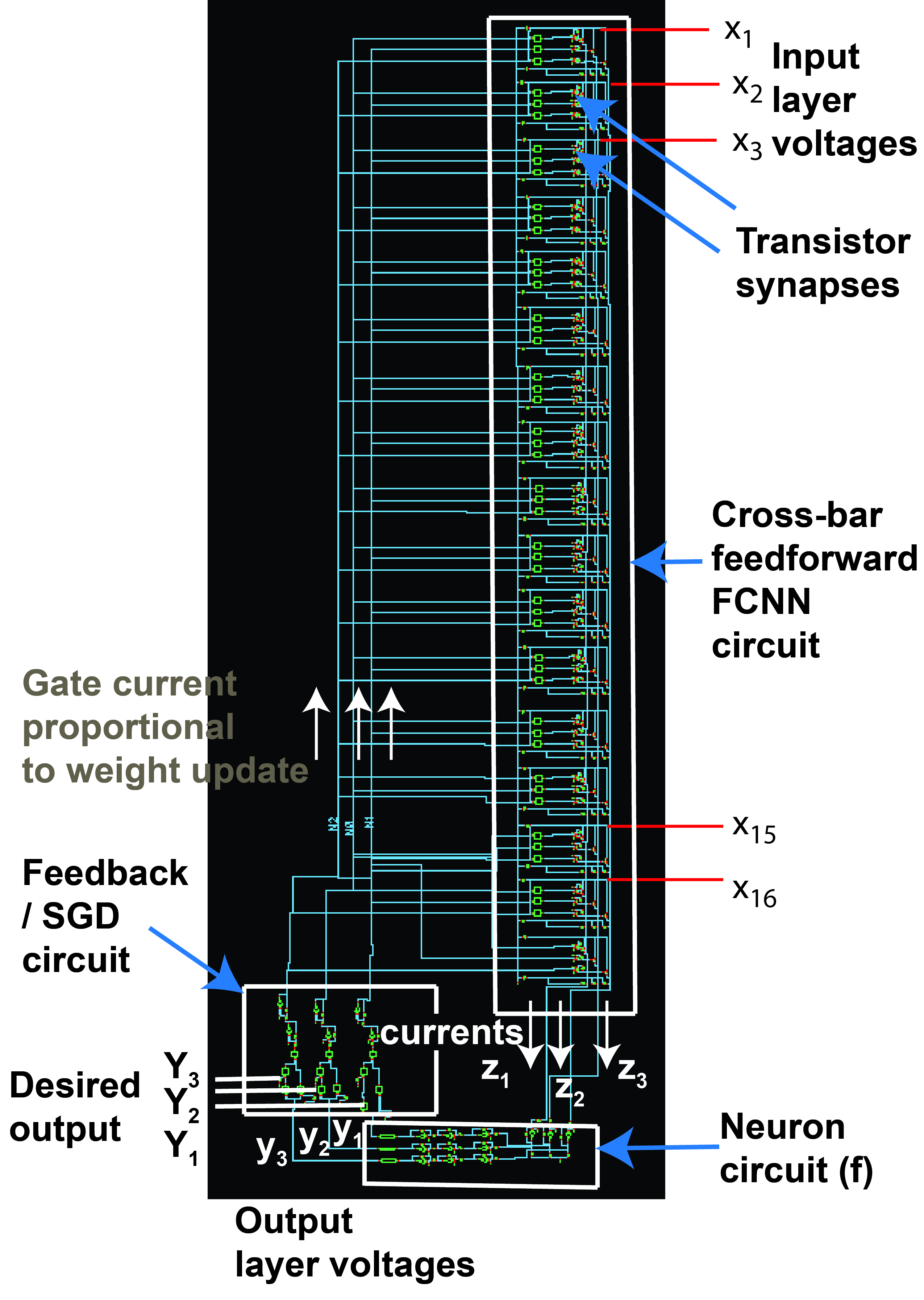}
  \caption{Schematic of crossbar circuit of Fig. ~\ref{Network1}, neuron circuit of Fig.~\ref{Neuron Circuit} at each output node of crossbar circuit and feedback circuit (SGD algorithm based) of Fig. ~\ref{SGD} at output node of crossbar circuit all wired together. The schematic is designed and simulated on Cadence Virtuoso (SPICE) circuit simulator for on-chip learning on Fisher's Iris dataset. }
  \label{Schematic}
\end{figure*} 

\begin{figure*}[!t]
  \centering
  \includegraphics[width=0.5\textwidth]{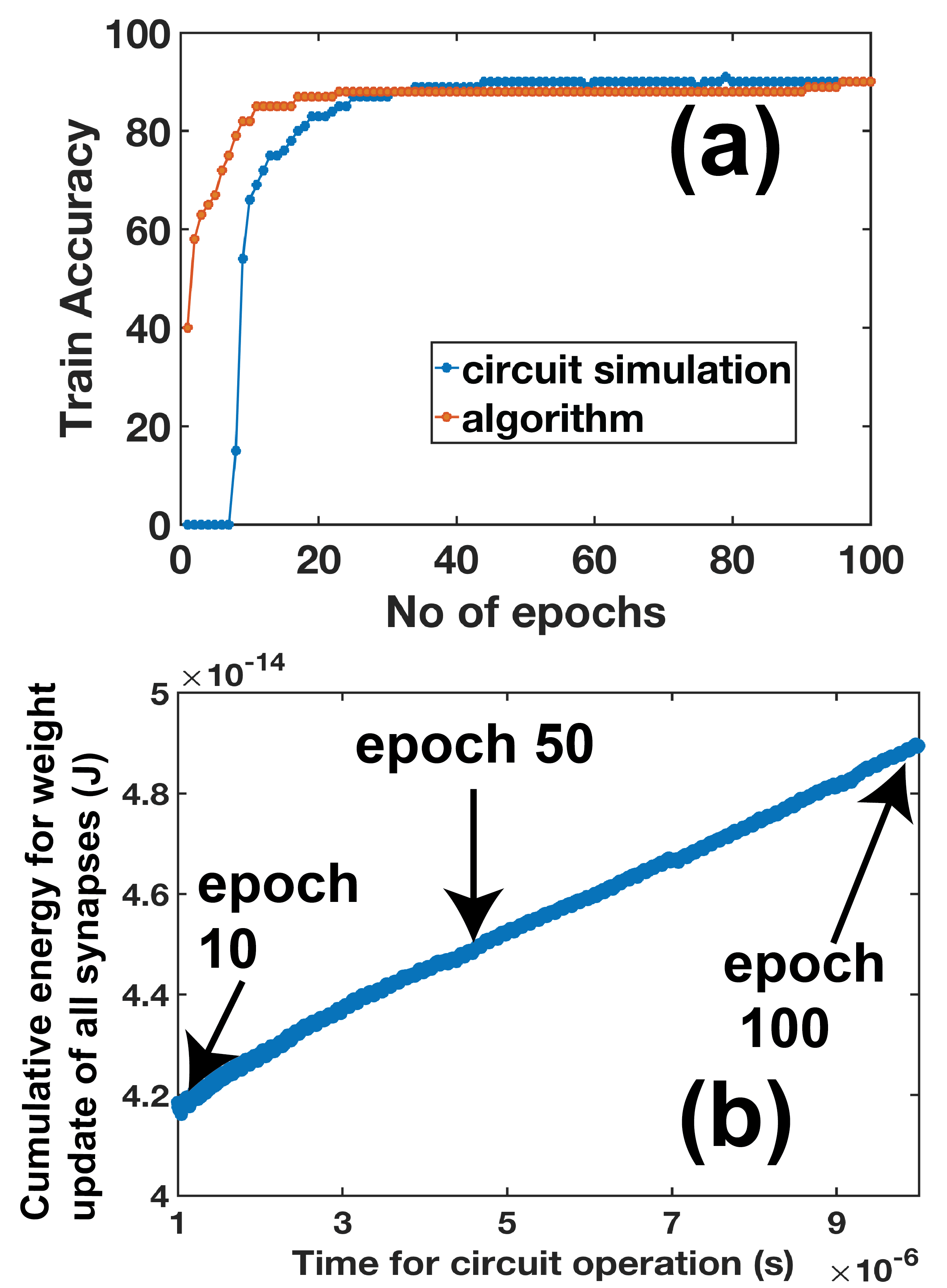}
  \caption{(a) Accuracy as a function of epochs on 100 train samples of Fisher's Iris dataset, obtained from SPICE simulations of circuit in Figure ~\ref{Schematic} (orange plot) and execution of the same algorithm on a conventional computer (software neural network). (b) Corresponding training energy (cumulative) in the designed circuit vs epoch is plotted. It includes energy consumption for weight update of all the synapses.}
  \label{Accuracy}
\end{figure*}

\begin{figure*}[!t]
  \centering
  \includegraphics[width=0.8\textwidth]{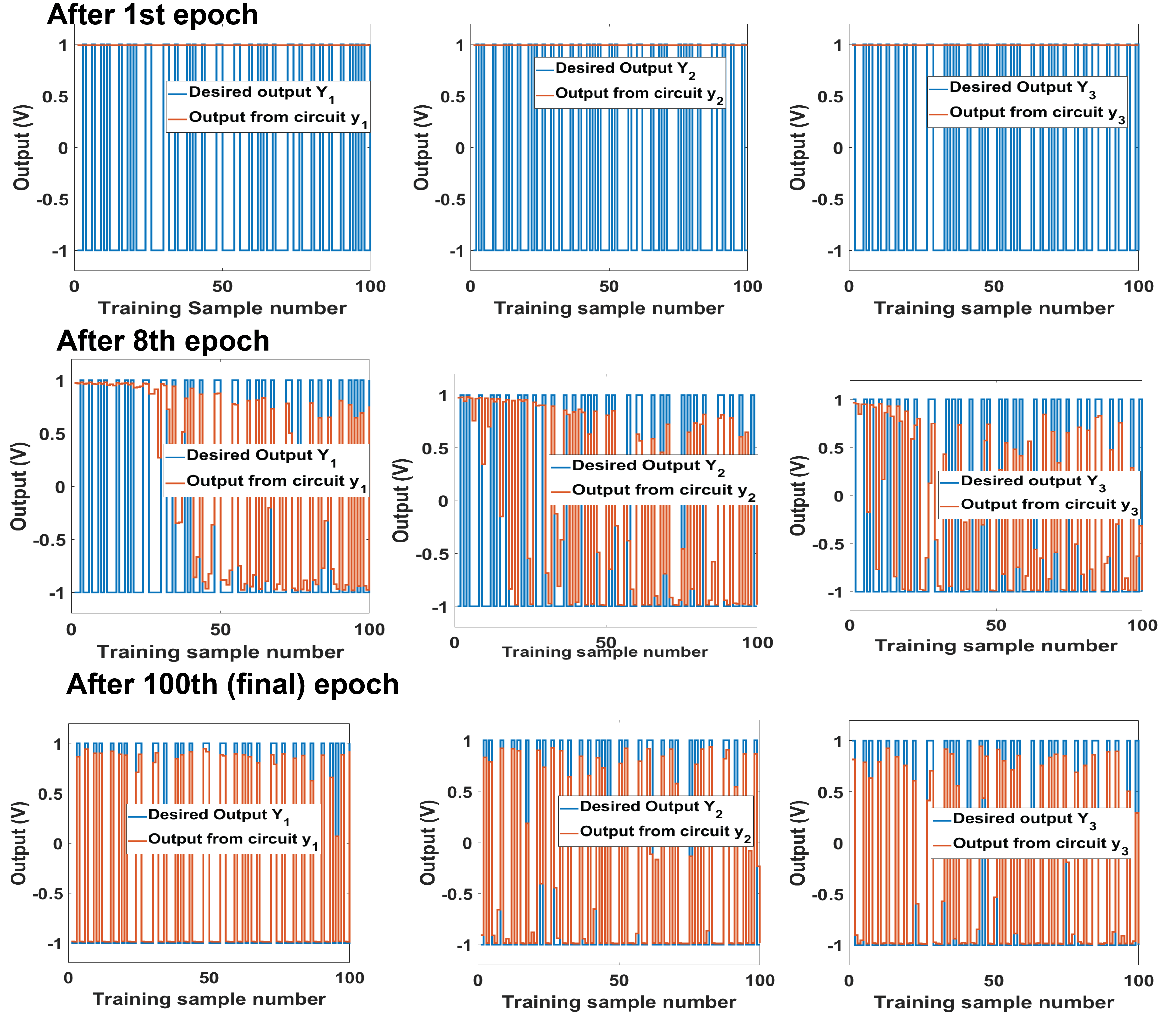}
  \caption{Output voltage of neuron circuit at each node of the circuit in Figure 6 , obtained from SPICE simulation, is plotted with respect to every training sample in an epoch. Results for 2nd epoch (a), 10th epoch (b) and 100th epoch (c) are shown.}
  \label{output}
\end{figure*}

Next this voltage is applied to a transistor based feedback circuit, also analog just like the neuron circuit. This circuit present at every output node updates the weights and trains the network in hardware, or rather accomplishes on-chip learning via  Stochastic Gradient Descent (SGD) calculation as discussed below (Fig.~\ref{SGD}).

 For almost every vector in the input set for which the training is done (training set) ${\{x_1,x_2,x_3...x_m..x_{M}\}}$ the output ${\{y_1,y_2,y_3...y_n..y_{N}\}}$ must match the desired output: ${\{Y_1,Y_2,Y_3...Y_n..Y_{N}\}}$, corresponding to the known label/ class that input vector belongs to. In order to do that, the weight matrix need to be updated adopting the Stochastic Gradient Descent (SGD) method through several iterations over the training set. 
 
 The error generated at output node $n$ is given by: 
\begin{equation}
\epsilon_{n}=\frac{1}{2}{(Y_n-y_n)^{2}}
\end{equation} 
Weight of the synapse connecting output node $n$ with input node $m$ ($w_{n,m}$) is updated between iteration $i$ and $i+1$ as follows:

\begin{multline}
w^{i+1}_{n,m}=w^{i}_{n,m}-\Delta w_{n,m}\\
=w^{i}_{n,m}-\eta\frac{\partial \epsilon_n}{\partial w_{n,m}}\\
=w^{i}_{n,m}-\frac{\eta\lambda}{2}(Y_n-y_n)(1-y^2_n)x_m
\end{multline}

and weight of the bias synapse for output node $n$ is updated as follows:

\begin{multline}
w^{i+1}_{n,0}=w^{i}_{n,0}-\Delta w_{n,0}\\
=w^{i}_{n,0}-\eta\frac{\partial \epsilon_n}{\partial w_{n,0}}\\
=w^{i}_{n,0}-\frac{\eta\lambda}{2}(Y_n-y_n)(1-y^2_n)
\end{multline}

where $\eta$ is the learning rate and each iteration corresponds to each training sample inside each epoch. Learning rate and the hyperparameter $\lambda$ can be adjusted in hardware, by changing the gain of an opamp, using a variable resistor in the "amplifier" block of the SGD calculation circuit (Fig.~\ref{SGD}(a)). The SGD circuit has been designed by us on Cadence Virtuoso using transistor and op-amp (made of transistors) based  subtractor and multiplier blocks as shown in Fig.~\ref{SGD}(b,c,d). The SGD circuit computes the weight update ($\Delta w_{n,m}$) \cite{Bhowmik_JMMM}. Building a subtractor block from op-amp is a standard process in analog electronics \cite{NeamenCircuit}. The multiplication operation is carried out with a single transistor making use of the fact that $I_{DS}$ is proportional to $V_{DS}$ times $V_{GS}$\cite{GeoffreyBurrIBMInternalReport}.

Since weight of each transistor synapse is proportional to the applied gate to source voltage ($V_{GS}$) (Fig.~\ref{Synapse1}(b)), in order to update the weight at each synapse, from equation (4) and (5) the gate voltage has to be updated as follows: 

\begin{multline}
V_{GS,n,m}^{i+1}=V_{GS,n,m}^{i}-\Delta V_{GS,n,m};\\
V_{GS,n,0}^{i+1}=V_{GS,n,0}^{i}-\Delta V_{GS,n,0}
\end{multline}

The weight update is calculated ($\Delta w_{n,m}$) by the SGD calculation circuit and generated in the form of a 1 ns long voltage pulse since input voltage pulse is 1ns long, corresponding to a training sample. A voltage controlled current source, working based on the principle of Howland current pump \cite{HowlandCurrentSource}, is designed at the output stage. It converts the voltage pulse to 1 ns long programming current pulse. Magnitude of current is proportional to $\Delta w_{n,m}$ (Fig.~\ref{SGD}(a),(d)).
 The current pulse is applied at the gate of the transistor such that change in gate voltage is proportional to the integral of gate current over time, as shown in the previous section, and is equal to the weight update of equation (6). Such weight update carried out over all transistor synapses over all training samples through several repetitions/ epochs results in on-chip learning of the designed FCNN.
 

\section{Accuracy, energy and speed analysis for on chip learning on Fisher's Iris dataset}

We next do SPICE simulations of the transistor synapse based crossbar FCNN circuit (Fig. ~\ref{Network1}), neuron circuit (Fig. ~\ref{Neuron Circuit}) and SGD based feedback/ weight update circuit (Fig. ~\ref{SGD}) we design, wired all together as shown in Fig. ~\ref{Network1}. The corresponding schematic of Fig. ~\ref{Network1} on the SPICE simulator (Cadence Virtuoso) is shown in Fig. ~\ref{Schematic}. We carry out on-chip learning for the circuit, following the method described in the previous section, on the Fisher's Iris dataset- a standard dataset in the machine learning community \cite{FisherIris}.

There are 16 input nodes in our FCNN circuit (Fig. ~\ref{Schematic}) corresponding to 16 inputs: 4 features of flowers passed through 4 sensors each. \cite{UdayanArchive}: 
$f(x) = x$,
$f(x) = 1-x$,
$f(x) = 1-2|x-0.5|$,
$f(x) = 2|x-0.5|. $ There are 3 output nodes corresponding to 3 possible classes of flowers. For  flower of type A, desired output ${Y_1,Y_2,Y_3}= {1,-1,-1}$.  For flower of type B, desired output ${Y_1,Y_2,Y_3}= {-1,1,-1}$. For flower of type C, desired output ${Y_1,Y_2,Y_3}= {-1,-1,1}$.  The available dataset has total 150 samples, almost equally distributed among types A,B and C. 100 samples are used here for training in each epoch and 50 separate samples are used for final testing to determine test accuracy. 

Each sample is trained for 1 ns.   Our transistor synapses can be operated at higher speed but magnitude of gate currents for weight updates and hence total energy for training will also go up. So we choose this speed of 1 ns per training sample in an epoch.Training accuracy vs epoch plot is shown in Fig. \ref{Accuracy} (a).  Accuracy number is obtained by comparing the voltage waveforms at the output nodes, i.e. at the output stage of the neuron circuits (${y_1,y_2,y_3}$), with the desired waveforms (${Y_1,Y_2,Y_3}$), from the SPICE simulation of our full neural network circuit shown in Fig.~\ref{Schematic} . For every sample, if the output at each of the three nodes is within 40 percent of the desired output then we consider it a success. For about the first five epochs the accuracy is 0. This is because all three output nodes ${y_1,y_2,y_3}$ are at {1,1,1} V for all samples. Hence the output is wrong for all the samples. Output for 2nd epoch for example is shown in Fig. ~\ref{output}(a). Around 10th epoch the output nodes start giving correct output and hence accuracy suddenly increases (Fig. ~\ref{Accuracy}(a)). Output waveform for 10th epoch is shown in Fig. ~\ref{output}(b). Beyond the 10th epoch, outputs at all nodes gradually start following the desired outputs. Thus accuracy increases with epoch as shown in Fig. ~\ref{Accuracy}(a). Output waveform for 100th epoch shows that all output nodes give the same output as desired for 90 samples. Thus after 100 epochs, or 100 x 100 iterations or 10 $\mu$s, the accuracy(both train and test) (on 100 train samples) is 90 percent which is very similar to algorithm implemented in python code.  Thus we  have been able to achieve successful on-chip learning of our transistor synapse based analog FCNN circuit in SPICE simulations.

\begin{table*}[!t]
  \caption{Comparison of performance for on-chip learning of our transistor synapse based FCNN circuit with implementation of the same FCNN using other synaptic devices on the same dataset (Fisher's Iris) }
  \label{tab:freq}
  \begin{tabular}{cccc}
    \toprule
    Type of synapse & Time per sample & Total time & Total energy consumed  \\
    & per epoch & for learning & in synapses while learning\\
    \midrule
    Transistor (this work)  & 1 ns & 10 $\mu$s & 50 fJ\\
    \midrule
    Domain wall based & 3 ns & 30 $\mu$s & 9 fJ \\
    spintronic device &  &  & \\
    \midrule
    Oxide based & between 200 ns (each pulse & variable & 1 $\mu$J\\
    RRAM device & for conductance increase) & & \\
    & and 6 $\mu$s ("Reset" pulse) & &\\
 \bottomrule
\end{tabular}
\end{table*}

From Fig. ~\ref{Accuracy}, by 50th epoch or 5 $\mu$s, train(and test) accuracy reaches 90 percent(b) and net energy consumed in the synapses is as low as $\approx$ 45 fJ. Energy consumed in 100 epochs is $\approx$ 50 fJ. It is to be noted that energy consumed in analog peripheral circuits is ignored in this calculation. Those circuits can be further optimized to enable ultra low energy on-chip learning on our proposed transistor synapses based analog hardware NN.

\section{Comparision of performance of proposed MOSFET synapse with other kinds of synaptic devices }

Next we compare the energy and speed performance of on-chip learning on our proposed MOSFET synapse based FCNNN with that on FCNN designed using some other kinds of synaptic devices that have been proposed and implemented elsewhere - spin orbit torque driven domain wall based synapses (spintronic synapses) \cite{Kaushik_IEEEReview, Bhowmik_JMMM, LongYouExp} and memristive oxide based RRAM synapses \cite{RRAMModel,RRAMNature}.  For fair comparison between the different synaptic devices, neural networks with the same architecture and number of nodes needs to be designed with different types of devices as synapses and they have to be trained on the same dataset using the same algorithm. Hence we simulate the same 16 input node x 3 output node FCNN circuit of Fig.~\ref{Schematic} with transistor synapses, domain wall synapses and RRAM based synapses. We use the same rule for weight update (Fig. ~\ref{SGD}) to train these three FCNN circuits.
Time for training each sample in each epoch (each iteration)  during on-chip learning, total time for the learning and energy consumed in the process are listed in Table 1 for the three FCNN circuits. The energy listed includes only energy consumed in the synapses for weight update during on-chip learning.

Our domain wall synapse model is micromagnetic in nature and is calibrated against experiments of current driven domain wall motion in Pt/Co/MgO devices \cite{PtCodomainwall1, PtCodomainwall2}. More details on our method for on-chip learning in domain wall synapse based FCNN circuit can be found in \cite{Bhowmik_JMMM}. Time for learning is comparable between transistor synapse FCNN and domain wall synapse FCNN circuit. Energy consumed is approximately 5 times higher for transistor synapses compared to domain wall synapses (Table I). However our transistor synapses are much easier to fabricate because they involve conventional silicon MOSFET, which is not the case with domain wall synapses which needs magnetic materials sputtered under specific conditions so that they exhibit perpendicular magnetic anisotropy and high Dzyoloshinski Moriya interaction \cite{SpintronicRoadmap, PtCodomainwall1, PtCodomainwall2}. 

Speed and energy performance of our designed transistor synapses based NN is also compared with similar NN designed using HfO$_x$ based RRAM synapse (Verilog model from \cite{RRAMModel} used in our simulation) with respect to on-chip learning on the same Fisher's Iris dataset in Table I. Though two RRAM devices are used per synapse in our simulation to address the asymmetry issue in conductance response \cite{PCMManan, GeffBurrTED, RRAMNature}, several long duration (6 $\mu$s) "Reset" pulses are still needed for successful weight update \cite{GeffBurrTED}. Hence a lot of energy is consumed for RRAM based NN compared to our transistor synapse based NN which doesn't have that issue, as explained earlier \cite{GeffBurrTED}. Also, since "Reset" pulse may or may not be needed for a particular synapse while training the NN on each sample in each epoch, the net time for training varies in the case of RRAM based NN in our weight update scheme. But time for training RRAM synapse based NN is certainly longer than our proposed transistor synapse based NN since even one "Reset" pulse is about a $\mu$s long while any pulse to increase or decrease weight in our transistor synapse is just 1 ns long, owing to its linear, bipolar synaptic characteristic (Fig. ~\ref{Synapse3}). Also, fabricating a RRAM based device requires in-house facilities and cannot be fabricated as easily as our proposed transistor synapse.

. 

\begin{figure*}[!t]
  \centering
  \includegraphics[width=0.6\textwidth]{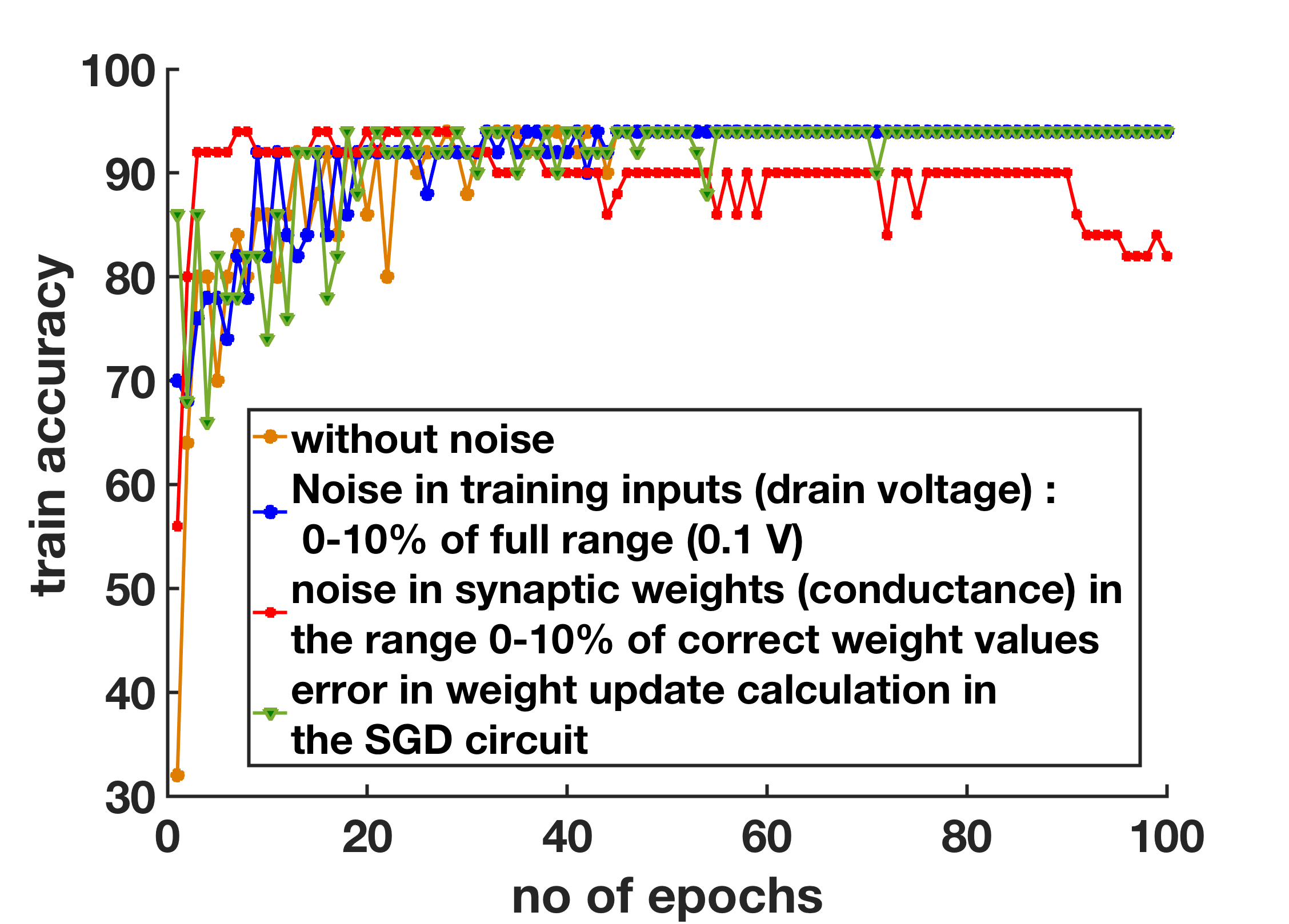}
  \caption{Training accuracy is plotted as function of epoch for the following four cases: without any noise in the circuit, in presence of noise in all training inputs through noise in drain voltages, in presence of noise in synaptic weights through device variation in conductances of transistor synapses and in presence of noise in SGD based weight update calculation in peripheral feedback circuit. }
  \label{Performance2}
\end{figure*}

\section{Device variability, noise in circuit voltages and retention of weights}

Since there can be device level variations among the different transistor synapses, their weights can vary from the values calculated by the weight update circuit. Considering noise in the range of 0-10 percent for all the synaptic weights, train accuracy still turns out to be almost the same as ideal transistor synapses from our calculation (Fig. ~\ref{Performance2}). This agrees with the observation in PCM synapse based FCNN, trained on-chip by similar gradient descent algorithm, that accuracy does not depend much on device stochasticity \cite{GeffBurrTED}. Similarly, we observe that noise in the training inputs through the drain voltages (Fig. ~\ref{Network1}) or error in weight update calculation up to 10 percent in the designed peripheral circuit does not affect the overall training accuracy much (Fig. ~\ref{Performance2}). 

 Though linear synaptic characteristic and easy, well developed method of fabrication are the two main advantages of our proposed transistor synapse compared to RRAM, PCM or spintronic synapses, unlike those Non Volatile Memory (NVM) based synapses, our proposed transistor based synapses do not retain their weights for a very long time. Our SPICE simulations show that once on-chip learning is achieved in the designed transistor synapse based NN circuit (Fig. ~\ref{Network1}) and training inputs are stopped from being fed, the synaptic weights decay in about 1 ms. This is because the weight is proportional to the conductance, which is in turn proportional to the gate voltage. As the gate oxide capacitance of the synapse discharges, gate voltage drops and hence the correct weight value is lost with time. However, the retention time is $10^6$ times higher than time for training each sample in an epoch (1 ns). For that retention factor, on-chip learning can still be achieved as argued in \cite{RPU1,RPU2} and seen in our simulations (Fig. ~\ref{Accuracy}). However this means that our proposed transistor synapse based NN is not suitable for applications where the training is done in a traditional computer and only testing/ inference is done in hardware (off-chip learning). On the other hand, in applications where training and testing both need to be done in hardware, our proposed system provides an easily fabricable platform for training in hardware (on-chip learning) as opposed to PCM or RRAM synapse  based NN (\cite{GeffBurrJPhysDReview}) and does not need high voltage for weight update like floating gate synapse based NN. For testing/inference after a certain time duration from training, the proposed solution is to store the final weights after training in a floating gate synapse based NN for testing purposes, where this needs to be done only once and thereby the endurance and energy consumption issues in writing weight values on floating gate synapses do not become major bottleneck \cite{FloatingGateDisadvantages, FloatingGateDisadvantages2}. 

\section{Conclusion}

Thus in this paper we have proposed a new functionality of conventional Si-SiO$_2$ based MOSFET as synaptic device in analog hardware NN. No floating gate is present in the synapse unlike previous proposals. Through SPICE simulations we demonstrate successful on-chip learning on Fisher's Iris dataset and compare its energy and speed performance with other implemenations of analog hardware FCNN with previously proposed spintronic and oxide based RRAM synapses.  Easy means of fabrication through established silicon based merchant foundries and linear, bipolar synaptic characteristics make our MOSFET synapse a potential candidate for implementation of analog hardware NN in the near future.



\section*{Acknowledgment}

The authors would like to thank Rajinder Singh Deol and Madhusudan Singh for help with the experimental measurement, and Apoorv Dankar, Anand Kumar Verma, Shouri Chatterjee and Ankesh Jain for help with the simulations.

\end{document}